\documentclass[pdflatex,sn-mathphys-num]{sn-jnl}

\usepackage{graphicx}%
\usepackage{multirow}%
\usepackage{amsmath,amssymb,amsfonts}%
\usepackage{amsthm}%
\usepackage{mathrsfs}%
\usepackage[title]{appendix}%
\usepackage{xcolor}%
\usepackage{textcomp}%
\usepackage{manyfoot}%
\usepackage{booktabs}%
\usepackage{algorithm}%
\usepackage{algorithmicx}%
\usepackage{algpseudocode}%
\usepackage{listings}%

\theoremstyle{thmstyleone}%
\theoremstyle{thmstyletwo}%

\theoremstyle{thmstylethree}%

\begin{document}

\title[Kolmogorov-Arnold Networks and Evolutionary Game Theory for More Personalized Cancer Treatment
]{Kolmogorov-Arnold Networks and Evolutionary Game Theory for More Personalized Cancer Treatment
}


\author{\fnm{Sepinoud} \sur{Azimi}}\email{s.azimirashti@tudelft.nl}

\author{\fnm{Louise} \sur{Spekking}}\email{l.m.spekking@tudelft.nl}

\author{\fnm{Kate\v{r}ina} \sur{Sta\v{n}kov\'a}}\email{k.stankova@tudelft.nl}

\affil{\orgdiv{Department of Engineering Systems and Services}, \orgname{Delft University of Technology},  \city{Delft}, \country{The Netherlands}}


\abstract{Personalized cancer treatment is revolutionizing oncology by leveraging precision medicine and advanced computational techniques to tailor therapies to individual patients. Despite its transformative potential, challenges such as limited generalizability, interpretability, and reproducibility of predictive models hinder its integration into clinical practice. Current methodologies often rely on black-box machine learning models, which, while accurate, lack the transparency needed for clinician trust and real-world application. This paper proposes the development of an innovative framework that bridges Kolmogorov-Arnold Networks (KANs) and Evolutionary Game Theory (EGT) to address these limitations. Inspired by the Kolmogorov-Arnold representation theorem, KANs offer interpretable, edge-based neural architectures capable of modeling complex biological systems with unprecedented adaptability. Their integration into the EGT framework enables dynamic modeling of cancer progression and treatment responses. By combining KAN's computational precision with EGT's mechanistic insights, this hybrid approach promises to enhance predictive accuracy, scalability, and clinical usability. }

\keywords{Kolmogorov-Arnold representation network, Evolutionary game theory, cancer, presonalized medicine}



\maketitle

\section{Introduction} \label{sec_intro}
Personalized cancer treatment represents a paradigm shift in modern medicine, transitioning from one-size-fits-all therapies to interventions tailored to the individual patient. This approach is particularly critical in oncology, where genetic, molecular, and environmental variations significantly influence disease progression and treatment outcomes. Advances in precision medicine enable oncologists to leverage genomic profiling and biomarker analysis to identify the most effective treatments for individual patients, reducing side effects and improving therapeutic efficacy. For instance, recent studies have demonstrated how personalized therapies based on tumor genomics enhance survival rates in aggressive cancers such as triple-negative breast cancer \cite{bianchini2022treatment}.

The benefits of personalized medicine extend beyond improved clinical outcomes. By avoiding ineffective treatments and preventing complications, personalized approaches also contribute to cost reduction in healthcare systems. Furthermore, the integration of cutting-edge technologies such as machine learning (ML) and molecular diagnostics paves the way a future where predictive analytics can guide precise treatment decisions. As cancer incidence continues to rise globally, the implementation of personalized strategies is becoming an essential element in advancing patient care and achieving sustainable health outcomes.

Despite its potential, personalized cancer treatment faces significant challenges. Predictive modeling, a cornerstone of precision oncology, has shown promise but is often limited by issues such as accuracy, generalizability, and transparency. For example, black-box ML models like Deep Neural Networks (DNN) provide high predictive accuracy but lack interpretability, leading to hesitancy among clinicians to trust their outputs. Furthermore, the lack of robust validation frameworks often undermines their applicability across diverse patient populations. Addressing these limitations is critical to unlocking the full potential of predictive modeling in personalized cancer treatment.\cite{topol2019high}. Generalizability is another pressing concern, as models trained on limited datasets frequently underperform when applied to diverse patient populations. A recent study highlighted that predictive models trained on specific genomic datasets performed poorly when validated on patients from different ethnic backgrounds \cite{liu2024variability}. Additionally, many algorithms face challenges in reproducibility and robust validation across independent datasets, raising concerns about their reliability in real-world clinical settings \cite{yu2018artificial}.

Ethical considerations further complicate the integration of these models into clinical workflows. Issues such as data privacy, algorithmic bias, and the need for regulatory approval present significant barriers to widespread adoption. Addressing these limitations is critical to unlocking the full potential of predictive modeling in personalized cancer treatment.

Evolutionary Game Theory (EGT) offers a promising avenue to address these challenges. By modeling the interactions between tumor cells, treatments, and the immune system as dynamic strategies, EGT provides insights into the evolutionary dynamics of cancer. It excels in optimizing adaptive therapies, such as timing and dosing regimens, by analyzing how cancer cell populations evolve under treatment pressures. However, EGT alone often falls short in terms of predictive accuracy and scalability, particularly for aggressive cancers with complex, nonlinear dynamics.

To enhance the applicability of EGT, this paper proposes developing a novel framework that integrates Kolmogorov-Arnold Networks (KANs) and EGT to address the limitations of existing predictive models. Inspired by the Kolmogorov-Arnold representation theorem, KANs introduce interpretable and adaptive machine learning models capable of modeling complex biological systems. Their extension into KAN-ODEs (Ordinary Differential Equations) provides a powerful tool for accurately describing dynamic systems such as cancer progression \cite{kan_ode}.

By exploring the strengths and weaknesses of KANs and EGT, this paper lays the groundwork for an innovative methodology that bridges theoretical and clinical oncology. The proposed integration enhances predictive accuracy and interpretability, establishing a new standard for addressing complex, adaptive systems in cancer treatment.

\section{Evolutionary Game Theory and Machine Learning: Advancing Cancer Treatment Amidst Constraints} \label{sec_big_picture}

Cancer research has long stood at the crossroads of biology, mathematics, and technology, leveraging diverse tools to decode the intricacies of tumor growth and resistance to treatment. Mathematical modeling and ML have emerged as pivotal players in this journey, each bringing unique strengths to the table. Mathematical models excel in offering mechanistic insights, enabling researchers to simulate tumor behavior under various conditions. Meanwhile, ML thrives on pattern recognition, uncovering subtle trends within complex datasets. However, both approaches face significant challenges: mathematical models often rely on oversimplified assumptions, while ML systems depend heavily on large, high-quality datasets. 

In this section we briefly present the contribution of these approaches to the field of oncology and some challenges hindering their wide adoption in clinical settings.

Mathematical oncology is an interdisciplinary field that uses mathematical modeling and optimization techniques to better understand and treat cancer~\cite{altrock2015,Kuang2016,west2023survey}. This can be done by creating models that simulate the behavior of tumors, predict how they will progress and respond to treatments, and optimize personalized therapies. Models are typically based on our understanding of cancer biology and are validated and / or inspired by real data, such as data from patients~\cite{ghaffari_laleh_classical_2022,soboleva2023validation,zhang2022evolution} or, for example, data from \textit{in vitro} experiments~\cite{rejniak2010current,kaznatcheev2019fibroblasts,Spekking2022.01.26.477755,werner2011dynamics}. We can use models to make recommendations so that oncologists can explain, test, or adjust their expectations or treatment strategies, and help them to make more effective, data-driven decisions to optimize patient outcomes.

Mathematical models are also used to optimize the timing and dosing of treatments~\cite{gallasch2013mathematical,chmielecki2011optimization,moore2018mathematically}. By taking into account elements such as drug resistance and the different cell types, the models can make patient-specific adaptive treatment predictions. 

More and more studies demonstrate the growing impact and importance of mathematical oncology~\cite{dujon2021identifying,LV_zhang2017integrating,RD_Gluzman2020optimizing,strobl2021turnover,powathil2013towards}. Game theory, a branch of mathematics, can improve cancer treatment by analyzing interactions between cancer cells of different types and framing the treatment process as a strategic interaction between the physician and the tumor~\cite{tomlinson1997game,archetti2019cooperation,Wolfl2020,orlando2012}. The latter approach focuses on understanding resistance dynamics and the relative sizes of different cell populations, rather than solely considering the tumor size~\cite{stavnkova2019optimizing,Garjani2023,blitz,westtowards}. By creating models that simulate tumor behavior, predict treatment responses, personalize therapies and allow a better understanding of cancer, mathematical oncology is a powerful tool in the fight against cancer~\cite{dujon2021identifying,RD_Kaznatcheev2017cancer,you2017spatial,RD_Gluzman2020optimizing,LV_Bayer2022coordination,LV_Grigorenko2022lotka,LV_zhang2017integrating}.

However, these models rely heavily on the adequacy of their underlying hypotheses and the quality of available data. The validity of their predictions often depends on whether the assumptions about tumor behavior and treatment responses accurately reflect reality. This dependency can limit their flexibility, especially in cases where biological systems are too complex or poorly understood.

ML on the other hand, bypasses the need for predefined hypotheses by learning patterns and relationships directly from the data. This data-driven approach allows ML models to uncover insights that might be inaccessible through traditional hypothesis-based method.

The integration of ML into cancer treatment has ushered in a new era of diagnostic and therapeutic capabilities, particularly through the use of black box models like DNNs. These models have demonstrated efficacy in predicting cancer outcomes and personalizing treatment plans. For instance, convolutional neural networks (CNNs) have been employed to analyze medical imaging data, achieving high accuracy in detecting lung cancer from low-dose computed tomography (CT) scans \cite{chao2021deep, mikhael2023sybil}. Furthermore, a study utilizing a multimodal data—integrating clinical, imaging, and molecular features— reported an enhanced predictive accuracy for breast cancer survival outcomes \cite{vale2021long}. The authors notably address the limitations of traditional models, such as Cox proportional hazards, by overcoming linearity and proportionality constraints, leading to more precise survival. Similarly, attention-gated convolutional neural networks (SiGaAtCNNs) have demonstrated superior feature extraction capabilities, yielding substantial improvements in classification accuracy and sensitivity for breast cancer prognosis \cite{arya2021multi}. Furthermore, deep learning models utilizing histopathology images without requiring pixel-level annotations have shown remarkable performance in stratifying risk and improving survival predictions across diverse cancer types \cite{wulczyn2020deep}. These advancements underscore the transformative potential of ML in oncology, offering precise prognostic insights and enhancing therapeutic decision-making  \cite{kalafi2019machine}. 

Black box models, as the name suggests, lack transparency, making their reliability a point of concern. Clinicians are often hesitant to adopt tools they cannot fully understand or explain, particularly in high-stakes environments like cancer treatment, where trust and accountability are paramount. Also despite these promising results, the development of explainable artificial intelligence (XAI) has not progressed as rapidly as anticipated. While researchers have made strides in creating models that provide some level of interpretability, the complexity of the algorithms often remains a barrier to widespread clinical adoption \citep{10.1155/2022/2965166}. For instance, while techniques such as Local Interpretable Model-agnostic Explanations (LIME) and SHapley Additive exPlanations (SHAP) have been proposed to elucidate model predictions, they do not fully address the need for inherent interpretability in clinical settings \citep{10.1038/s41467-023-40890-x}. As a result, there is a growing consensus that the field requires models that are interpretable by design, particularly in cancer treatment, where the stakes are high and the consequences of misinterpretation can be severe \citep{10.1021/acs.jcim.0c00331}. By prioritizing the development of interpretable models, the healthcare sector can better leverage the capabilities of ML while ensuring that clinicians and patients can trust the outputs of these advanced technologies.  In addition to the challenges of interpretability, the integration of ML models into clinical workflows is also hindered by the need for robust validation and regulatory approval. The clinical environment demands not only high accuracy but also reliability and safety in the deployment of ML systems. For instance, a study highlighted that while ML models could accurately predict patient outcomes, the inability to explain how these predictions were made led to hesitance among clinicians to rely on them \citep{10.3322/caac.21708}. This skepticism is further exacerbated by the potential for false positives, which can have serious implications for patient care \citep{10.11591/eei.v13i2.6280}. 

By developing models that not only excel in predictive accuracy but also offer clear, interpretable outputs, the healthcare sector can better leverage the power of ML while ensuring patient safety and fostering clinician confidence in these advanced technologies \citep{10.1186/s13662-019-2324-9}.  The need for interpretable models is particularly pressing in the context of cancer treatment, where the complexity of the disease and the variability in patient responses necessitate a nuanced understanding of treatment effects. 

\section{Kolmogorov-Arnold Networks: Bridging the Gap} \label{sec_kan_gap}
\begin{figure}[h!]
    \centering
    \includegraphics[width=\textwidth]{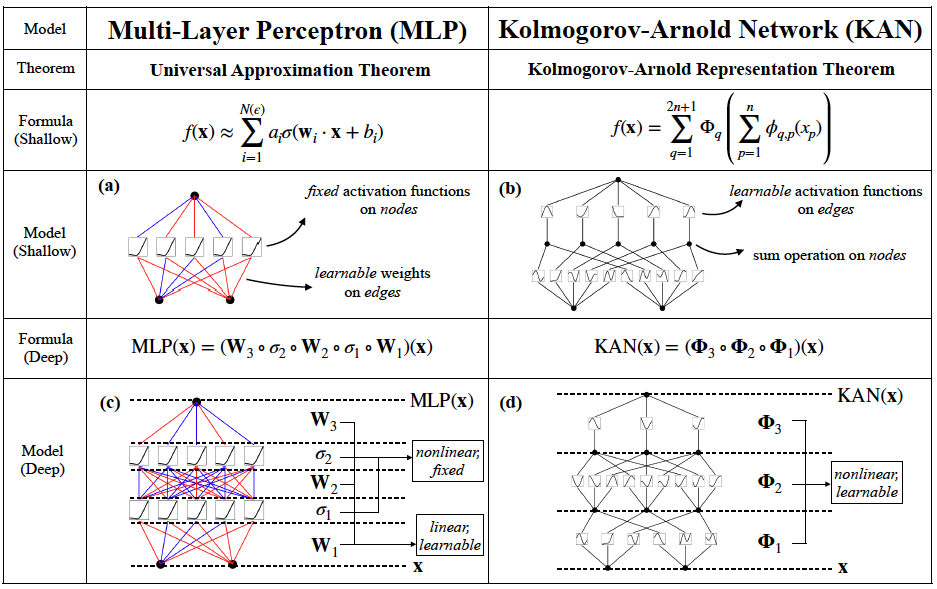}
    \caption{Multi-Layer Perceptrons (MLPs) vs. Kolmogorov-Arnold Networks (KANs), Figure from \cite{kan_survey2}.}
    \label{fig:kan_mlp}
\end{figure}

As the demand for high-performing yet interpretable ML models continues to grow, KANs have emerged as a transformative solution. Unlike traditional neural networks such as Multi-Layer Perceptrons (MLPs), as is depicted in \cite{kan_original} and presented in Figure \ref{fig:kan_mlp}, which rely on fixed activation functions at nodes, KANs redefine the neural architecture by placing learnable univariate functions on edges.

This design innovation is inspired by the Kolmogorov-Arnold representation theorem, which states that any continuous multivariate function $f(\mathbf{x})$ can be expressed as a finite sum of continuous univariate functions, mathematically represented as:

\begin{equation}
    f(\mathbf{x}) = \sum_{q=1}^{2n+1} \Phi_q \left( \sum_{p=1}^n \phi_{q,p}(x_p) \right),
\end{equation}

where $\phi_{q,p}(x_p)$ and $\Phi_q$ are learnable univariate functions. This structural shift from node-based to edge-based learning allows KANs to provide interpretable and adaptive representations of high-dimensional data, making them highly efficient in solving complex computational problems \cite{kan_original, kan_survey1, kan_survey2}.

KANs fundamentally change the way neural networks learn and represent data. Each edge in a KAN represents a learnable function, offering flexibility in the architecture to model intricate relationships. The forward propagation in a KAN layer is given by:

\begin{equation}
    \mathbf{x}_{l+1, j} = \sum_{i=1}^{n_l} \phi_{l, j, i}(\mathbf{x}_{l, i}),
\end{equation}

and the network output can be represented as a hierarchical composition of these functions:

\begin{equation}
    \text{KAN}(\mathbf{x}) = \Phi_{L-1} \circ \Phi_{L-2} \circ \cdots \circ \Phi_0 (\mathbf{x}),
\end{equation}

where $\Phi_l$ is the function matrix at layer $l$. This mathematical formulation not only highlights the novelty of KANs but also underscores their ability to adapt to diverse and complex data structures \cite{kan_original}.

\begin{figure}[h!]
    \centering
    \includegraphics[width=\textwidth]{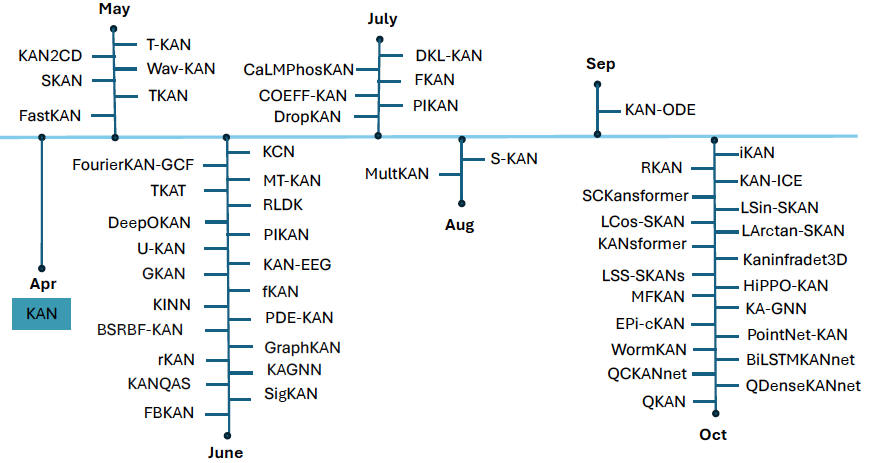}
    \caption{Progression of KAN advancements in 2024, Figure from \cite{kan_survey2}.}
    \label{fig:kan_progression}
\end{figure}

The introduction of KANs has generated remarkable excitement in the field of machine learning, evidenced by the publication of comprehensive surveys exploring their variations and applications \cite{kan_survey1, kan_survey2}. These surveys highlight the rapid evolution of KANs, from their foundational architecture to specialized variants designed for specific use cases. For example, Temporal-KAN (T-KAN) \cite{xu2024kolmogorov} extends KANs for sequential data analysis by integrating memory mechanisms akin to recurrent neural networks. T-KAN excels in time-series forecasting, outperforming traditional RNNs by efficiently modeling long-term dependencies. Another pivotal variant, Wavelet-KAN (Wav-KAN) \cite{bozorgasl2024wav}, employs wavelet-based activation functions to perform multi-resolution analysis, enhancing noise robustness and interpretability, particularly in high-dimensional and noisy environments \cite{kan_survey1, kan_survey2}. A comprehensive overview of how KAN's development has progressed in 2024 is depicted in \cite{kan_survey2}, see Figure \ref{fig:kan_progression}.

Among these advancements, KAN-ODE \cite{kan_ode} stands out as a promising extension, bridging the gap between machine learning and mathematical modeling. While the original KAN framework demonstrated its capability in solving partial differential equations (PDEs), KAN-ODE further advances this by integrating KANs into the Neural Ordinary Differential Equation (Neural ODE) framework. This integration allows the model to learn and infer the dynamics of complex systems directly from data. The KAN-ODE framework is governed by:

\begin{equation}
    \frac{du}{dt} = \text{KAN}(u(t), \theta),
\end{equation}

where $u$ represents the state variables and $\theta$ denotes the trainable parameters of the network. This formulation provides an interpretable and efficient mechanism for uncovering the hidden physics of dynamical systems, such as wave propagation and phase separation, even in scenarios with sparse training data \cite{kan_ode}.

\begin{figure}[h!]
    \centering
    \includegraphics[width=\textwidth]{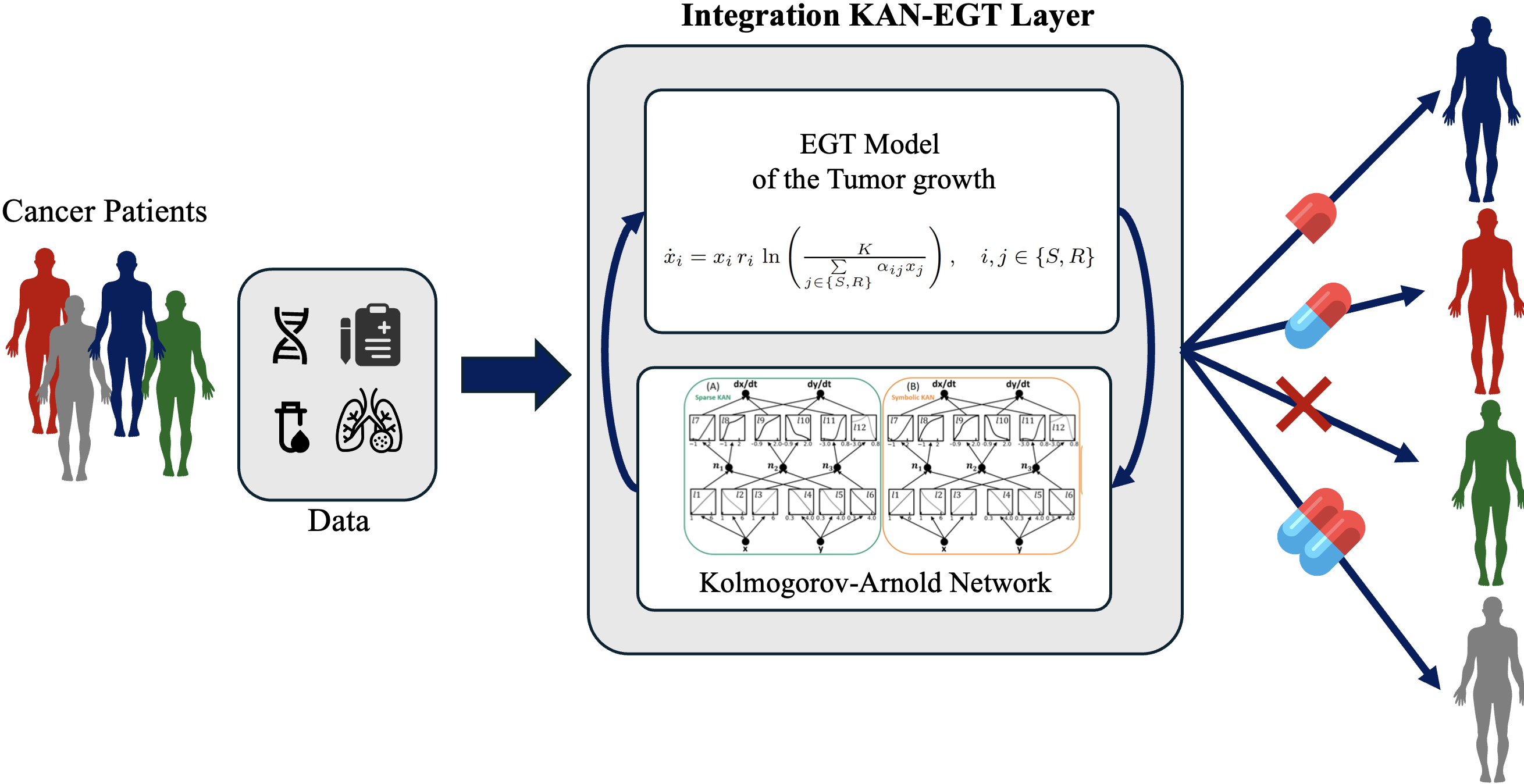}
    \caption{Kolmogorov-Arnold Networks (KANs) and Evolutionary Game Theory (EGT) Modeling intergradation for better personalized cancer treatment, KAN Figure from \cite{kan_ode}.}
    \label{fig:integration}
\end{figure}

The potential of KAN-ODE in personalized cancer treatment, as depicted in Figure \ref{fig:integration}, is particularly promising. EGT has been successful in optimizing medication strategies for non-aggressive cancers, yet it has faced limitations in addressing more aggressive forms like non-small cell lung cancer (NSCLC). By integrating KAN-ODE with EGT, this approach combines mechanistic insights with the predictive power of machine learning, enabling the identification of personalized, optimal treatment strategies. This integration not only has the potential to offer interpretable models of cancer progression but also uncovers hidden mechanisms driving resistance and facilitates the exploration of novel interactions and adaptive responses without prior hypotheses \cite{kan_ode, kan_survey2}.

The synergy between KANs and EGT holds the promise to create a unified framework that merges the realism of EGT with the predictive capabilities of machine learning. Such a framework can dynamically optimize treatment strategies, offering clinicians actionable insights into the evolutionary dynamics of cancer and providing a pathway toward more effective and personalized therapies. By uniting interpretability with computational power, KANs represent a monumental step forward in the quest to bridge theory and application, offering transformative possibilities in scientific discovery and medical innovation.

\section{Challenges and Research Directions} \label{sec_challenges_future}
The integration of KAN and EGT represents an innovative approach in leveraging machine learning and mathematical mdoelling for personalized medicine. Yet, several challenges remain in ensuring its practical application and scalability. KAN is still in its early stages, with most existing studies limited to small-scale experiments. The lack of extensive validation across diverse clinical datasets raises concerns about its robustness and generalizability. Whether KAN’s theoretical advantages can translate into consistent clinical outcomes is a critical question that time and research must answer.

One of the foundational issues lies in the original Kolmogorov-Arnold Theorem (KAT), which relies on stringent smoothness conditions to guarantee function representation. While KAN claims to mitigate these constraints by employing learnable univariate functions, there is limited empirical evidence to validate this claim. The absence of formal guarantees on the stability and adaptability of these functions in complex, noisy datasets poses another limitation. Moreover, for KAN to achieve its full potential, high-resolution, longitudinal datasets capturing detailed patient-specific information are crucial. However, such datasets are often challenging to obtain due to privacy concerns, inconsistent data collection practices, and the high costs associated with large-scale, standardized data acquisition.

The integration of Neural Ordinary Differential Equations (Neural ODEs) within the KAN framework amplifies its computational complexity. Training KANs on real-world cancer datasets, characterized by high dimensionality and sparse features, demands substantial computational resources. This poses a barrier to widespread adoption, particularly in resource-constrained settings. Beyond computational challenges, there is also a persistent gap between algorithmic development and clinical implementation. Integrating KAN-based models into clinical workflows requires interdisciplinary collaboration among data scientists, oncologists, and healthcare policymakers. Addressing regulatory and ethical concerns surrounding the deployment of AI in medicine is equally critical to achieving real-world impact.

Future research must prioritize the development of optimized, scalable versions of KAN to address the computational demands. Leveraging techniques such as model pruning, parallel computing, and efficient training algorithms could significantly reduce the resource burden. Establishing robust validation frameworks is equally essential to bridge the gap between theory and practice. These frameworks should encompass diverse cancer datasets, including multi-modal data integrating clinical, genomic, and imaging features, to assess KAN’s generalizability and reliability. Additionally, incorporating interpretability into the KAN framework is a priority. Developing intrinsic explainability mechanisms, such as feature attribution maps for univariate functions, could enhance clinician trust and facilitate regulatory approval. Aligning KAN’s outputs with clinical decision-making processes will expedite its acceptance in real-world settings.

 Beside KAN's exploration in oncology, the application to other complex, multi-agent systems—such as neurodegenerative diseases, autoimmune disorders, and other similarly challenging fields—represents an intriguing avenue for future research. Expanding its use cases can unlock broader opportunities for personalized medicine. The synergy between KAN and EGT also presents an exciting avenue for optimizing treatment strategies. Future research should explore deeper integration, leveraging KAN’s interpretability to model and predict adaptive responses in multi-agent systems, including tumor-immune interactions.

\section{Conclusion} \label{sec_conclusion}
The integration of KANs with EGT provides a novel conceptual framework for addressing the complexities of adaptive treatment strategies in multi-agent systems. KANs, with their edge-based, learnable architecture, offer an interpretable and adaptable method for modeling high-dimensional biological systems. Coupled with the dynamic modeling capabilities of EGT, this framework has the potential to improve the optimization of treatment regimens and to provide insights into disease progression and therapeutic resistance. This position paper outlines the theoretical foundation for this framework, serving as a starting point for future research rather than presenting a fully developed implementation.

The proposed framework requires further investigation to address existing challenges. Computational demands, particularly in the integration of Neural ODEs, need to be mitigated through advancements in techniques such as model pruning and parallelization. Collaboration among data scientists, clinicians, and policymakers will also be essential to align the framework with clinical and practical requirements, ensuring that it is both robust and applicable.

Although the focus here is on oncology, the framework may also be applicable to other complex systems, including neurodegenerative diseases and autoimmune disorders. These conditions, with their multi-agent interactions and variability in patient responses, present additional opportunities to explore the potential of the KAN-EGT approach. Future studies should aim to extend and validate this framework in these areas, contributing to its development as a generalizable tool for personalized medicine.
\bibliography{sn-bibliography}

\end{document}